# Deep Image Compositing

Shivangi Aneja
Technical University of Munich,Germany
shivangi.aneja@tum.de

Soham Mazumder
Technical University of Munich,Germany
soham.mazumder@tum.de

## ABSTRACT

In image editing, the most common task is pasting objects from one image to the other and then eventually adjusting the manifestation of the foreground object with the background object. This task is called image compositing. But image compositing is a challenging problem which requires professional editing skills and a considerable amount of time. Not only these professionals are expensive to hire, but the tools (like Adobe Photoshop) used for doing such tasks are also expensive to purchase making the overall task of image compositing difficult for people without this skillset. In this work we aim to cater to this problem by making composite images look realistic. To achieve this, we are using Generative Adversarial Networks (GANS). By training the network with a diverse range of filters applied to the images and special loss functions, the model is able to decode the color histogram of foreground and background part of the image and also learns to blend the foreground object with the background. The hue and saturation values of the image plays an important role as discussed in this paper. To the best of our knowledge, this is the first work that uses GANs for that task of image compositing. Currently, there is no benchmark dataset available for image compositing. So we created the dataset and will also make the dataset publicly available for benchmarking. Experimental results on this dataset show that our method outperforms all current state-of-the-art methods.

## 1. INTRODUCTION

Image Compositing is the most common task in image editing. To create a composite image, the foreground object from one image is pasted on the background of another image. Most of the time, appearances (like color, lighting and texture compatibility) of foreground objects are inconsistent with the background on which it is pasted on since they are from different images, making the composite image unrealistic. Therefore, it is essential to adjust the appearances of the foreground object to make it compatible with the new background. In this work, we propose a conditional GAN [1] for the task of image compositing. Our generator is a simple autoencoder network with skip links that takes unrealistic composite images as input makes it realistic. Our discriminator is inspired from PatchGAN [2], which instead of predicting that image is real or fake, outputs a patch and outputting real or fake for every pixel in the patch, thus acting as a strong regularizer and penalizing more during training. The novelty of our work lies in training the network with a variety of special losses discussed in subsequent sections that helps the network nicely blend the foreground object with the background object. The contribution of our work is two fold. First, to the best of our knowledge this is the first work that uses GANS for the task of image compositing and our work eventually surpasses the current state-of-the-art [3]. Secondly, since there is no available dataset for image compositing benchmarking we created the dataset and would be making it publicly available for others in order to benchmark their results. The dataset used, network architecture, loss functions and results are described in subsequent sections.

## 2. RELATED WORKS

Most recent deep learning approaches for image compositing namely RealismCNN [4] and Deep Image Harmonization [3] improve the realism of composite images but have certain limitations as discussed below. [4] first learns a model that discriminates between real and composite images using a simple binary classifier and predicts the degree of realism in the image. It then uses this learnt model as a guidance tool to improve the quality of composite images. They learn a color adjustment model by optimizing for the realism score given by the classifier trained above and a regularization loss by constraining it to lie only in space of possible adjustments. They aim to penalize large change between the original object and recolored object, and prevent independent color channel variations (roughly hue change). [3] learns an end-to-end deep convolutional neural network (CNN)for harmonization of foreground parts with background. It takes as input a composite image and corresponding foreground mask, and outputs a harmonized image, where the appearances of the foreground region is adjusted with the background. Their deep CNN model consists of an encoder to capture the context of the input image and two decoders. One decoder to reconstruct the harmonized image using the learned representations from the encoder. Another decoder to provide scene parsing of the input image, while sharing the same encoder for learning feature representations. These techniques give good results when a filter similar to background is applied to the foreground object, but fails to give impressive results when the filter applied to foreground object is very different from the background or when appearances of the foreground and background regions are vastly different, leading to unreliable results. Also, in [4] first we learn a classifier and then use this to learn another model that adjusts colors. This requires a lot of GPU compute power. [3] uses a single network for training (encoder with two decoders) but along with image it requires image segmentation mask which is not always available and requires a lot of human-annotation effort. Our network improves the quality of composite images but with minimum compute power and without using image segmentation masks. Our work overcomes the limitation of current state-of-the-art, thus enabling us to generate realistic images in minimum time and without any human-annotation effort.

## 3. DATASET

One of the most essential aspects of training a deep convolutional network is data. GANs are generally trained on datasets of image pairs. The dataset needs to have enough variance so that it can learn to generalize on unseen data. For our task of realistic compositing, we need a dataset with image pairs of raw composites which are visibly different, and colour blended composites which look realistic. The raw composite image serves as the input to the generator network, while the blended composite image is the ground truth. Unlike unsupervised methods like [5], which can easily obtain training pairs, image compositing requires expertise to generate an image with high quality realism. This is not attainable for large-scale training data. So to tackle this problem, we take a real image as the ground truth. Then we proceed to edit the foreground of the real image, such that the edited image has a mismatched background and foreground colour palette. This edited image is taken as input to the network. Thus we acquire our training image pairs. This dataset building process ensures that the ground truth is a real image. So, the proposed network will strive to construct a realistic image given a raw composite input. The following section explains the process of achieving the composite inputs.

### 3.1 Synthesis of Composites

To assemble the image pairs for our task, we needed a segmentation dataset which contained masks of the foreground objects. The iCoSeg Dataset [6] serves our purpose. This dataset contains images with simple foreground objects, as well as the segmentation mask of the corresponding object. An example pair from the iCoSeg [6] is depicted in Figure 1.

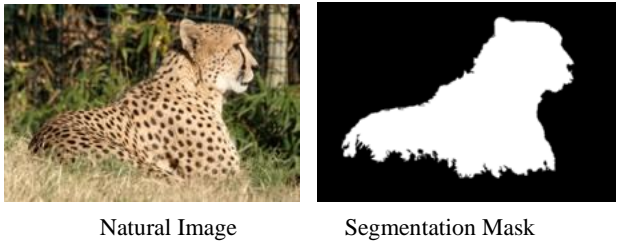

Natural Image        Segmentation Mask

**Figure 1. Segmenting the foreground object**

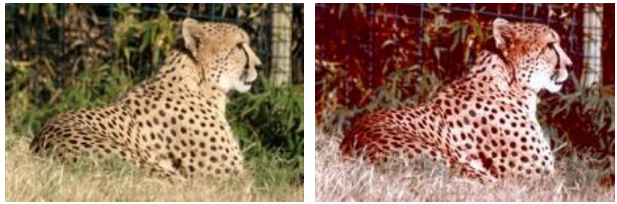

Natural Image        Stylised Image

**Figure 2. Applying styles to foreground object**

We use natural images from iCoseg[6] as the ground truth for our task. To generate the edited images we make use of the method presented in [7]. At first, we select a set of images whose colour style we wish to transfer to the background of the ground truth. Then using [7], we apply the styles to the natural images to obtain the edited images, as shown in Figure 2. We term them, stylised images. The second part of our dataset creation process involves editing the stylised image to get our desired raw composite inputs. To this end, the segmentation masks are used to cut out the foreground from the stylized images and subsequently paste it onto the ground truth natural images. The result is our Composite Image, which will serve as our input. This entire process is explained in Figure 3.

Using this method, we create a dataset of **3750** training and **370** test images.

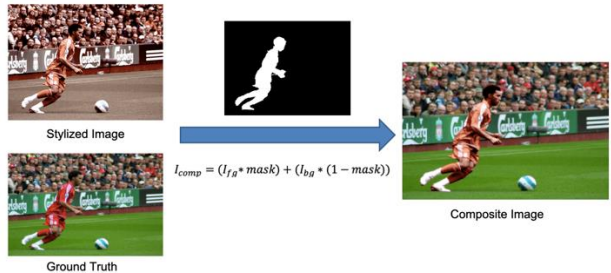

**Figure 3. Image compositing pipeline**

### 3.2 Discussion

The aforementioned dataset construction method provides a large collection of high-quality training image pairs. Additionally, this will ensure that the ground truth images are actual real images, so the network can really learn the representation of natural images and adjust the input image accordingly. Another advantage of our pipeline is the possibility of quantitative evaluations. We are now able to compare the synthesised composites to the ground truth images, using various image similarity metrics. This gives us a sense of how the images generated by our network fair against a real natural image (discussed in Section 6).

## 4. NETWORK ARCHITECTURE

We decided to use a Generative Adversarial Network [8] to solve our task. The motivation behind choosing a GAN, is its innate ability to learn feature representation. Our idea was to use a GAN to learn the colour distribution of real images, which would enable it to adjust the unnatural colour palette of the inputs and reconstruct very realistic composites. In a traditional GAN[8], we have no way to control the modes of data to be generated. However for our task, we need to generate images which preserve the structure of our composite inputs. This is achieved using Conditional GAN [1]. In conditional GAN we add the ground truth as an additional parameter to the generator to ensure structural similarity of the generated images. The ground truth images are also added to the discriminator input to distinguish real images better.

### 4.1 Generator

The generator network follows a symmetric autoencoder architecture. A single encoder block consists of a convolution with $3 \times 3$ filter dimension, followed by Leaky ReLU [9] activation and Batch Normalization [10] layer. We use a relatively small network with four such encoder blocks followed by four similar decoder blocks (which has transpose convolution instead). Also, we add skip links between the corresponding encoder and decoder blocks. These skip connections help to preserve image details and texture which are otherwise lost during convolutions. Furthermore, we remove the bottleneck layer of the autoencoder to further enhance the quality and reduce blurriness of the generated image.

## 4.2 Discriminator

We experiment with two types of architecture for the discriminator. The first one is a deep CNN with two residual blocks [11]. This discriminator maps from an image to a single scalar output which signifies the image is real or fake. The second one is inspired from the PatchGAN used in Pix2Pix [2]. We use four custom convolution layers proposed in their paper for the discriminator. Each convolution is followed by Batch Normalization [10] and then Leaky ReLU [9]. The PatchGAN discriminator on the other hand, outputs a smaller patch of the input image which is either real or fake. The entire architecture of our network is shown in Figure 4.

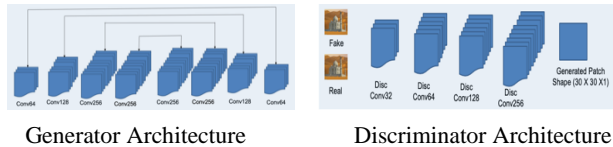

Generator Architecture      Discriminator Architecture

**Figure 4. Network architecture**

## 5. LOSS FUNCTIONS
### 5.1 HSV Loss

This loss is based on Hue, Saturation and Intensity values of an image. During experiments we found that the hue and saturation values of the image played an important role in preserving the realism of the image. So, we computed a novel loss function in the HSV colour representation. We first convert RGB images to HSV colour space and then compute channel wise L2 loss for the Hue and Saturation channels only. This loss is given by the following equation.

$$L_{hsv} = L_{hue} + L_{sat}$$

We presume that the colour palette of an image is easier to learn for the network when the image is in HSV representation. Further, only the Hue and Saturation dimensions are necessary to learn a realistic colour distribution. The idea behind this function is simple, yet quite effective for our task. To the best of our knowledge, no other method for image compositing uses this.

### 5.2 Reconstruction Loss
For reconstruction we can either use L1-loss or L2-loss on the RGB images. We used pixel-wise L1-loss for our network.

### 5.3 Perceptual Loss
This loss function was proposed in [12] for real-time style transfer between the pair of images. This loss is based not on differences between pixels but instead on differences between high-level image feature representations extracted from pre-trained convolutional neural networks. It measures image similarities more robustly than per-pixel losses. Optimizing the network with this loss preserves the overall structure of the image and prevents distortion of the generated image.

### 5.4 Adversarial Loss
To train a GAN, we need an end to end loss function. It is called adversarial loss. It is a min-max loss function where the two networks are jointly trained with opposite goals. The generator tries to fool the discriminator, so it is trained to maximise the final classification error (between true and generated data). On the other hand, the discriminator tries to detect fake generated data, so it is trained to minimise the final classification error.

The Generator loss is the combination of losses, where $\lambda$'s are weights assigned to each loss during training. These weights are hyperparameters of the network that we tune to get the optimal performance. The generator loss is given below.

$$L_{generator} = L_{adversarial} + \lambda_1 L_{recon} + \lambda_2 L_{perceptual} + \lambda_3 L_{hsv}$$

The discriminator is trained with only adversarial loss, given by the following equation.

$$L_{discriminator} = L_{adversarial}$$

## 6. METRICS
Metrics are measures of quantitative assessment, which are used to determine the performance of a model and how the model performs when compared to other methodologies. The choice of metrics is very important. It will influence how the performance of machine learning algorithms is measured and compared. This section briefly sheds light on the metrics that we used to quantitatively compare our results for evaluation.

### 6.1 Mean Squared Error
This is an absolute difference between generated image and ground truth image. A low value indicates that the generated image is closer to ground truth. This is not a perfect measure to evaluate the realism of composite images.

### 6.2 Peak Signal To Noise Ratio (PSNR)
This is the ratio of maximum possible signal power to the power of corrupting noise. This metric is commonly used to measure the quality of reconstruction in image compression. This will give an idea of how much information is lost in the predicted images relative to ground truth images.

### 6.3 Structural Similarity Index (SSIM)
This metric predicts the perceived quality of an image using luminance masking and contrast masking. It gives an idea how structurally similar two images are, 0 being the minimum (no similarity) and 1 being the maximum (perfect similarity). For a realistic image, this value should be high.

### 6.4 Visual Information Fidelity (VIF)
This metric interprets image quality as "fidelity" with the reference image, the higher the better. It is deployed in the core of the Netflix video quality monitoring system, which controls the picture quality of all encoded videos streamed by Netflix.

## 7. RESULTS
We ran our experiments with two types of discriminator, keeping the generator network constant. We did an extensive hyperparameter search for weights of each loss function and learning rates of generator and discriminator networks. The optimal combination of the parameters we found is shown in Table 1. We discovered through our experiments that the PatchGAN based discriminator architecture is able to capture the realism aspect much better compared to the discriminator with resnet blocks. Here we show a simple example in Figure 5. Evidently, the PatchGAN based model generates a more realistic image.

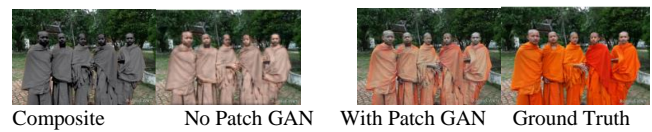

Composite     No Patch GAN     With Patch GAN     Ground Truth

**Figure 5. Comparison of different discriminator networks**

Our best architecture therefore, is with the PatchGAN discriminator. We present our results of this architecture and then compare with Zhu et al.[4] and the current state of the art Tsai et al.[3].

The qualitative results are shown in Figure 6. Visually, our method improves over Tsai et al.[3] (state of the art). Images generated by our model look more realistic and are visually closer to the ground truth. To further validate our results, we analysed the performance of our model quantitatively. We use the metrics (from Section 6) to rate our performance. Table 2 shows our findings. It is evident that our model outperforms Zhu et al. [4] as well as the state of the art Tsai et al.[3].

Our method has truly learnt to generate realistic composite images. This is backed by the findings shown in Figure 7. As can be seen, our model doesn't overfit on the dataset. Rather, it tries to balance the colour palette of the foreground with the background. It modifies the unnatural colour of the foreground to fit the realistic colour representation it has learnt through training.

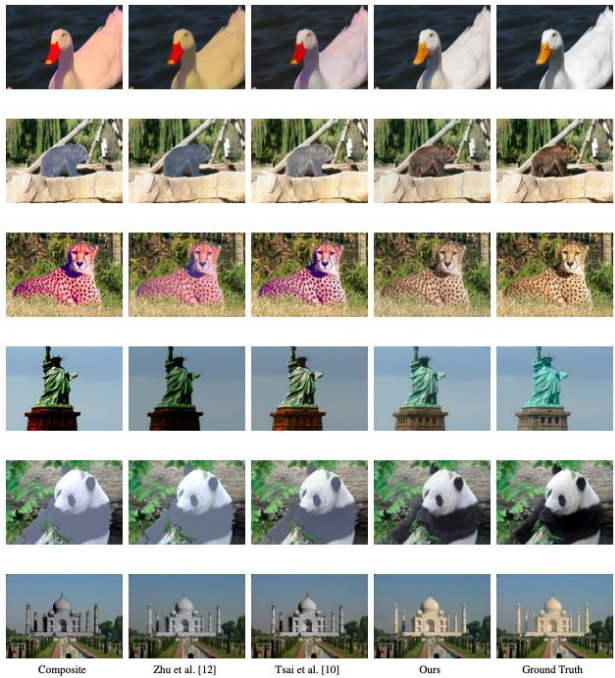

**Figure 6. Comparison of results on synthesized dataset**

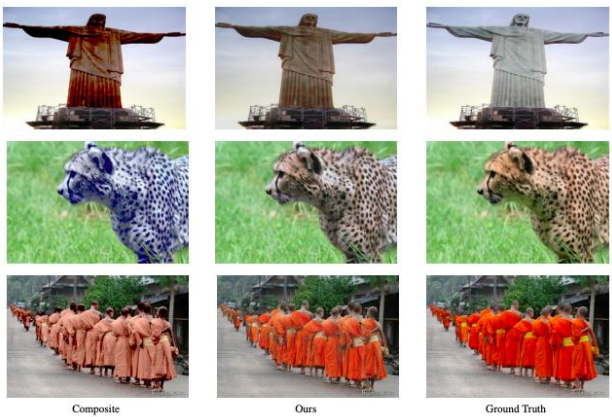

**Figure 7. Generated images which are realistic but not similar to ground truth.**

## 8. CONCLUSION

In this work, we use GAN to generate images with a high degree of realism. Training the generator with HSV loss helps the network to learn the color distribution of natural images and makes them more realistic. We conclude that our method with PatchGAN discriminator performs significantly better than the current state-of-the art. The patch based discriminator network significantly improve the realism in generated images. Results in Figure 7 show that our network is able to learn a diverse range of filters and generates images that appear visually realistic. This analysis is further supported by quantitative assessment in Table 2. A very important aspect of our network is that we don't use the foreground mask during training. The model learns to detect and correct the colour on its own. This is another major advantage of our network over [3].